\documentclass{article}

\usepackage{arxiv}
\usepackage[utf8]{inputenc} 
\usepackage[T1]{fontenc}    
\usepackage{url}            
\usepackage{booktabs}       
\usepackage{amsfonts}       
\usepackage{nicefrac}       
\usepackage{microtype}      
\usepackage{graphicx}
\usepackage[numbers]{natbib}
\usepackage{doi}
\usepackage{multirow}
\usepackage{amsmath}
\usepackage{graphicx}
\usepackage{caption}
\usepackage{hyperxmp}
\usepackage{csquotes}
\usepackage{makecell}

\title{Experimentation in Content Moderation using RWKV}
\author{
\begin{tabular}{c@{\hspace{8em}}c}
  \bf Umut YILDIRIM & \bf Rohan Dutta \\
   Highlight US Inc. & Meta Platforms, Inc. \\
  \texttt{umut@highlight.ing} & \texttt{rohandutta@g.ucla.edu} \\
  \\
  \bf Burak YILDIRIM & \bf Atharva Vaidya \\
  Gebze Technical University & Highlight US Inc. \\
  \texttt{b.yildirim2019@gtu.edu.tr} & \texttt{atharva@highlight.ing} \\
\end{tabular}
}

\usepackage{hyperref}

\makeatletter%
  \hypersetup{%
    pdfinfo={%
    Title={\@title},%
    Author={Umut YILDIRIM, Rohan Dutta, Burak YILDIRIM, Atharva Vaidya},%
    Subject={cs.AI, cs.CL, cs.CV, cs.MM},%
    Producer={pdfTeX},%
    Creator={Umut YILDIRIM, Rohan Dutta, Burak YILDIRIM, Atharva Vaidya},%
    Keywords={Content Moderation, Multimodal Language Models, RWKV, LLaVA, Multimodal Dataset, Multimedia Content},%
    },%
    keeppdfinfo 
}
\makeatother%

\renewcommand{\headeright}{}

\begin{document}
\maketitle
\begin{abstract}
This paper investigates the RWKV model's efficacy in content moderation through targeted experimentation. We introduce a novel dataset specifically designed for distillation into smaller models, enhancing content moderation practices. This comprehensive dataset encompasses images, videos, sounds, and text data that present societal challenges. Leveraging advanced Large Language Models (LLMs), we generated an extensive set of responses—558,958 for text and 83,625 for images—to train and refine content moderation systems. Our core experimentation involved fine-tuning the RWKV model, capitalizing on its CPU-efficient architecture to address large-scale content moderation tasks. By highlighting the dataset's potential for knowledge distillation, this study not only demonstrates RWKV's capability in improving the accuracy and efficiency of content moderation systems but also paves the way for developing more compact, resource-efficient models in this domain. Datasets and models can be found in HuggingFace: \hyperlink{https://huggingface.co/modrwkv}{https://huggingface.co/modrwkv}
\end{abstract}

\keywords{Content Moderation \and Multimodal Language Models \and RWKV \and LLaVA \and Multimodal Dataset \and Multimedia Content}
\section{Introduction}
\label{sec:intro}

Today's digital realm demands robust content moderation to preserve the integrity and ensure the safety of online social platforms. This safeguard is essential, shielding users from offensive or dangerous material, such as hate speech [\citealp[]{schmidt2017survey},\citealp[]{davidson2017automated}], graphic imagery [\citealp[]{duncombe2020social}], deceptive information [\citealp[]{nakamura2019r},\citealp[]{vosoughi2018spread}], and other violations. Content moderation underpins the enforcement of platform-specific norms and standards, crucial for fostering user confidence and protecting the platform's brand. It also mitigates legal liabilities by proactively addressing potentially unlawful content like defamation or copyright breaches.[\citealp[]{EliLilly2023}] 

Moreover, moderation plays an indispensable role in cultivating a civil and productive online environment. It discourages hostile conduct and promotes interactions that are both respectful and enriching. Such an environment is fundamental not only to user satisfaction and retention but also to maintaining advertiser appeal and upholding investor confidence.[\citealp[]{ForbesXAdSafety2023}, \citealp[]{EliLillyTwitter2023}]

The ever-evolving landscape of social media thrives on the rich tapestry of user-generated content that spans texts, images, videos, and sounds. Each mode of communication offers unique ways for users to share thoughts, experiences, and creative works, fostering a vibrant space for dialogue and cultural exchange.

Despite this, content moderation practices lag behind, often ill-equipped to fully grasp the nuanced and layered nature of user conversations.[\citealp[]{HBRArticle2022ContentModerationIsTerriblebyDesign}, \citealp[]{schmidt-wiegand-2017-survey}] It's within this context that the need for Large Language Models (LLMs) becomes apparent. LLMs are critical in discerning the complexities of online discourse, where meaning is frequently interwoven across different types of media. By leveraging a comprehensive dataset that encapsulates this multimodal reality, our project seeks to enhance content moderation, targeting the multifaceted and sometimes hidden pernicious content. Such an advanced multimodal strategy is essential to effectively filter out harmful material and promote a healthy, respectful digital environment that nurtures constructive interactions.[\citealp[]{chen2023combating}] 
\section{Related Work}

\subsection{Rule By Example: Harnessing Logical Rules for Explainable Hate Speech Detection}

The paper [\citealp[]{clarke2023rule}] introduces Rule By Example (RBE), a novel exemplar-based contrastive learning approach for textual content moderation, aiming to combine the transparency and interpretability of rule-based methods with the predictive power of deep learning models. Classic approaches to content moderation rely on rule-based heuristics [\citealp[]{baly2018integrating},\citealp[]{das2020detecting}], which, while transparent and easy to interpret, lack flexibility and robustness. Deep learning models have shown promise in overcoming these limitations but often lack transparency, leading to mistrust and adoption challenges. RBE addresses this by learning rich embedding representations for hateful content and logical rules governing them, allowing for explainable predictions via rule-grounding. Experimental results demonstrate that RBE outperforms state-of-the-art classifiers in both supervised and unsupervised settings on three benchmark datasets, showcasing its effectiveness and potential for improving content moderation systems.

\subsection{Rethinking Multimodal Content Moderation from an Asymmetric Angle with
Mixed-modality}

The paper [\citealp[]{yuan2024rethinking}]  introduces Asymmetric Mixed-Modal Moderation (AM3), a novel content moderation model designed to address the challenges of moderating multimodal content on social media platforms. Traditional unimodal moderation systems may struggle to detect harmful content that spans multiple modalities, such as memes combining images and text. AM3 features a unique asymmetric fusion architecture that preserves the distinct characteristics of each modality while effectively combining their information. To address the semantic asymmetry between vision and language, AM3 employs a cross-modality contrastive loss to learn the unique knowledge conveyed by multimodal content. Extensive experiments demonstrate that AM3 outperforms existing methods on both multimodal and unimodal content moderation tasks, showcasing its effectiveness in handling diverse types of harmful content online.
\section{Dataset and Data Preparation}

We have developed a comprehensive multimodal instruction dataset for content moderation, encompassing text, images, audio, and video. Our dataset incorporates content from various publicly available sources:
\begin{table}[h!]
\centering
\def\arraystretch{1.5}
\begin{tabular}{|l|l|l|}
\hline
\textbf{Modality} & \textbf{Sources} & \textbf{Content Type} \\
\hline
Text & \makecell{Civil Comments [\citealp[]{DBLP:journals/corr/abs-1903-04561}], OIG Moderation [\citealp[]{ontocord}], \\ OpenAI moderation [\citealp[]{openai2024gpt4}]} & User posts, conversations \\ \hline
Image & \makecell{LSPD [\citealp[]{Phan2022LSPDAL}], NSFW GitHub repo [\citealp[]{ebazarov_nsfw}], \\ Kaggle Violence Images [\citealp[]{karandeep98_real_life_violence}]} & NSFW, violence, normal images \\ \hline
Audio & Real Life Violence Situations Dataset [\citealp[]{mohamedmustafa_violence_situations}] & Extracted from videos \\ \hline
Video & LSPD [\citealp[]{Phan2022LSPDAL}], VSD [\citealp[]{demarty2014vsd}], NDPI2k [\citealp[]{moreira2016pornography}, ], XD-Violence \citealp[]{mohamedmustafa_violence_situations}] & Violence, pornography, normal videos \\
\hline
\end{tabular}
\caption{Overview of Data Sources}
\label{tab:data_sources}
\end{table}

\subsection{Data Preparation Process}

Our data preparation process involves several key steps:

1. \textbf{Data Collection}: We aggregated data from multiple sources (Table \ref{tab:data_sources}) to ensure diversity in content and scenarios.

2. \textbf{Preprocessing}: 
   - Text: Cleaned and normalized user posts and conversations.
   - Images: Filtered and categorized based on content type.
   - Audio: Extracted from videos and transcribed using OpenAI's Whisper\cite{radford2022robustspeechrecognitionlargescale}.
   - Video: Extracted frames and audio tracks for multi-modal analysis.

3. \textbf{Instruction Generation}: We used GPT-4 to create a set of 20 diverse instructions for each modality, focusing on content moderation tasks.

4. \textbf{Instruction-Response Pair Creation}: 
   - Text: Used GPT4 model to generate responses to instructions based on input text.
   - Images: Employed GPT4V to generate descriptions and moderation decisions.
   - Audio/Video: Combined transcriptions, visual analysis, and moderation instructions.

5. \textbf{Multimodal Integration}: Aligned data from different modalities to create a cohesive dataset that captures the complexity of real-world content moderation scenarios.

\begin{figure}[h!]
  \centering
  \includegraphics[width=0.9\linewidth]{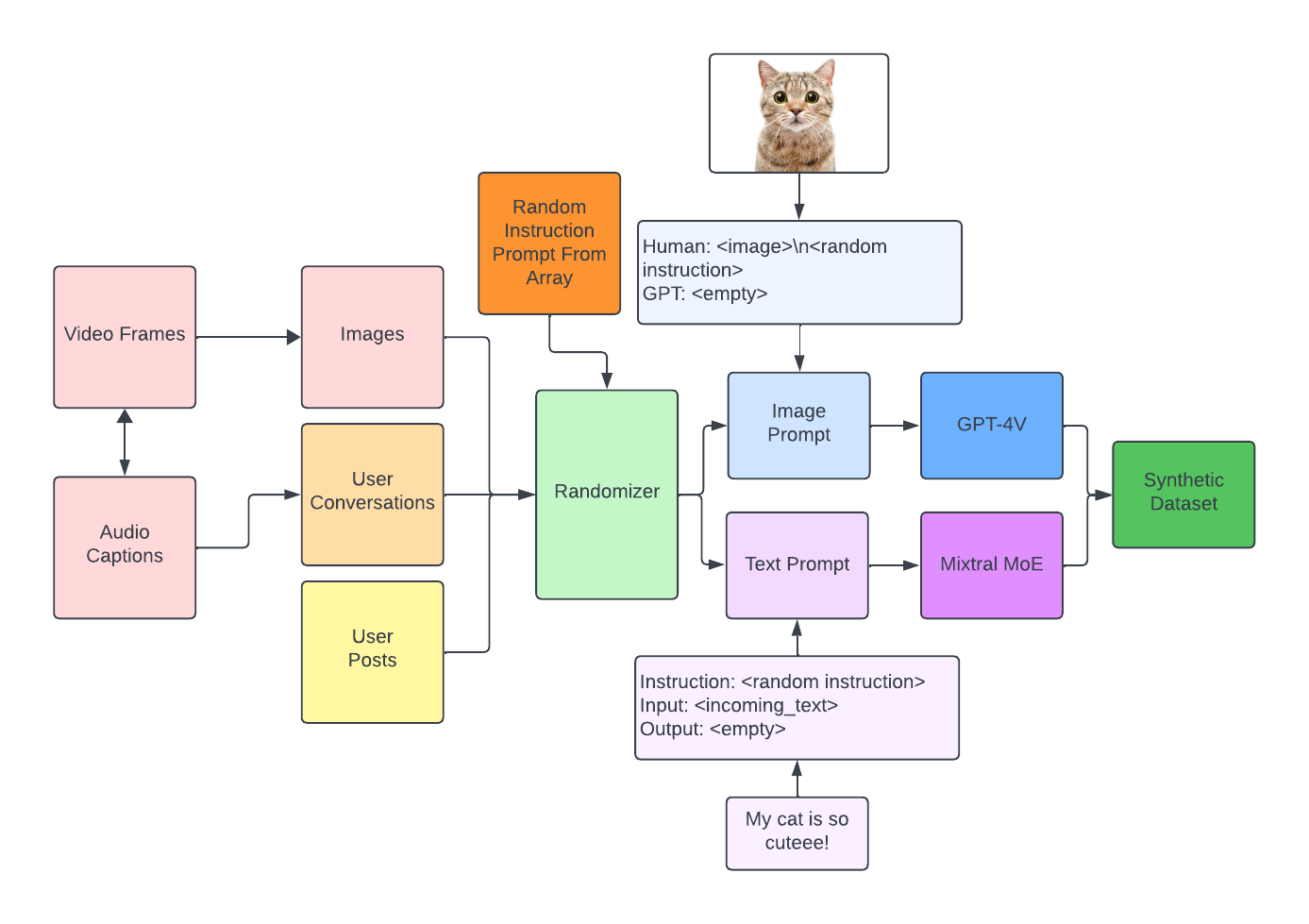}
  \caption{Synthetic dataset generation process}
  \label{fig:syntheticdatagen}
\end{figure}

\begin{figure}[h!]
    \centering
    \includegraphics[width=0.5\linewidth]{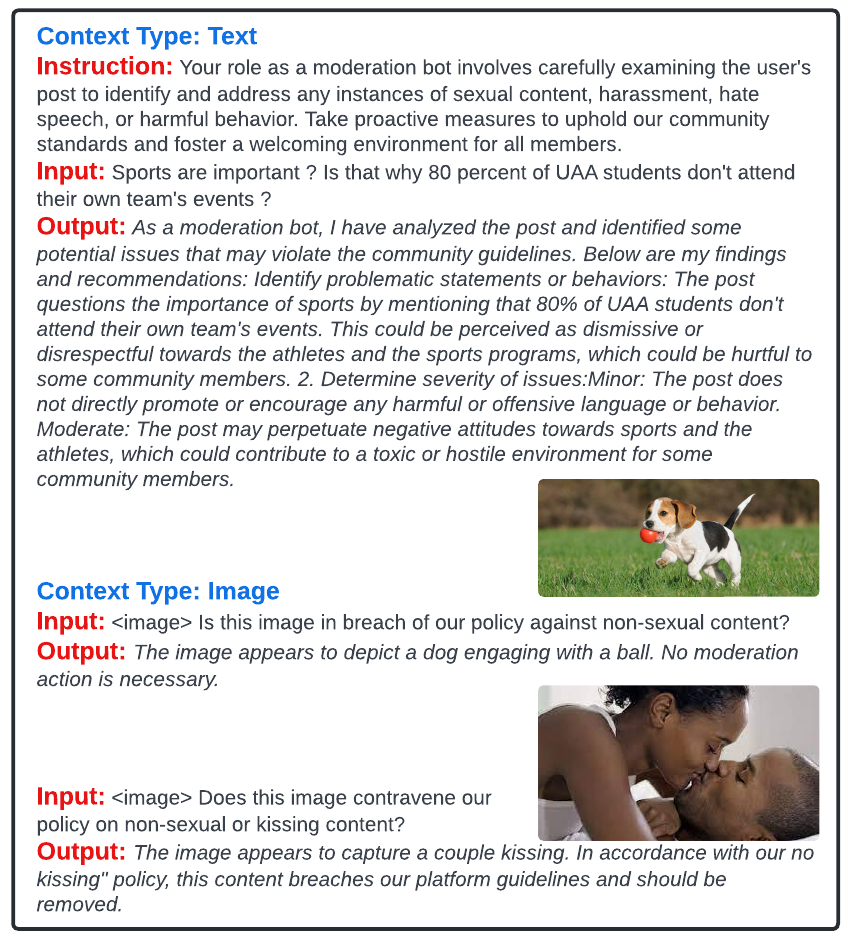}
    \caption{Example conversation with Mod-RWKV and Mod-VisualRWKV}
    \label{fig:exampleconversationwithrwkvandvisualrwkv}
\end{figure}

\subsection{Instruction-Response Pair Examples}

To illustrate our approach, here are examples of instruction-response pairs for different modalities:

\begin{table}[h]
\centering
\def\arraystretch{1.3}
\begin{tabular}{|l|l|p{6cm}|}
\hline
\textbf{Modality} & \textbf{Instruction} & \textbf{Response} \\
\hline
Text & Evaluate this comment for hate speech & Input: \enquote{All [ethnic group] should leave the country!} \\
& & Response: This comment contains hate speech targeting a specific ethnic group. It should be flagged for moderation... \\
\hline
Image & Identify any NSFW content in this image & [Description of image content] \\
& & Response: The image contains explicit nudity and should be classified as NSFW... \\
\hline
Audio & Detect signs of violence in this audio clip & [Transcription of audio] \\
& & Response: The audio contains sounds of shouting and physical altercation, indicating potential violence... \\
\hline
\end{tabular}
\caption{Example Instruction-Response Pairs}
\label{tab:instruction_examples}
\end{table}

\begin{figure}[h!]
  \centering
  \includegraphics[height=0.35\linewidth]{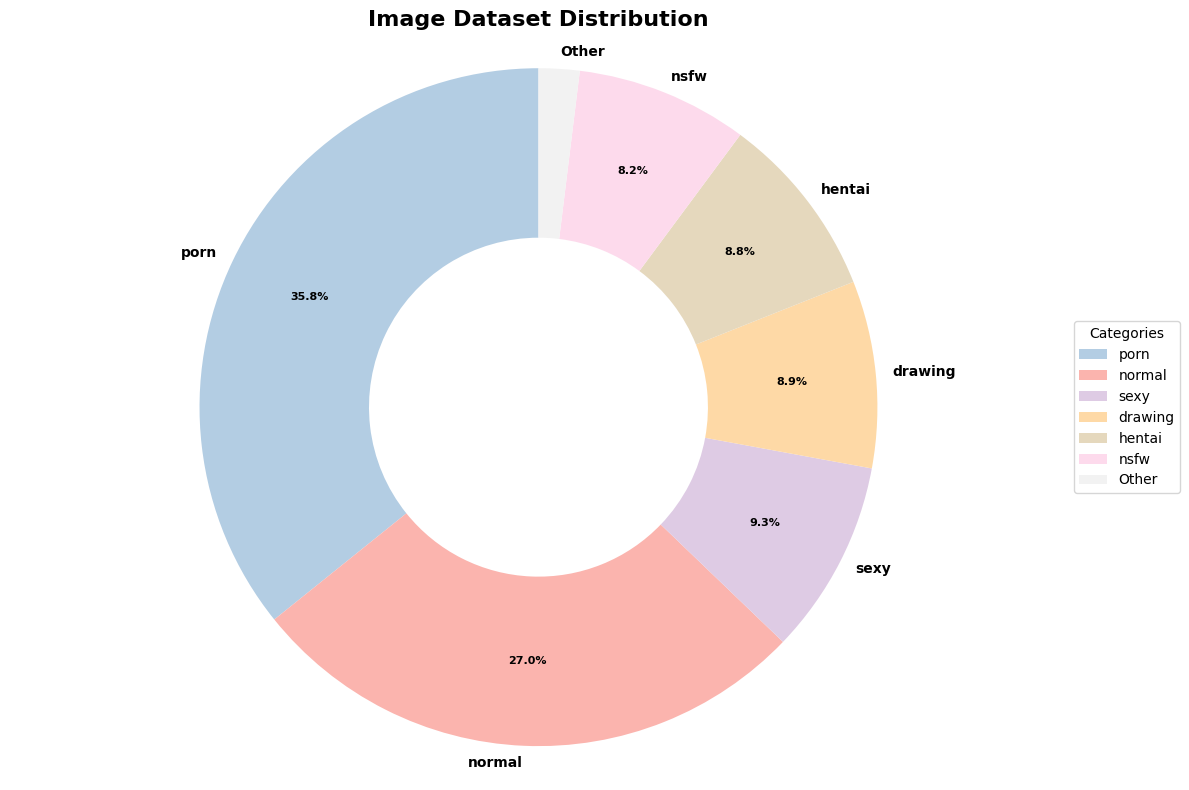}
  \caption{Image Dataset Distribution}
  \label{fig:imagedist}
\end{figure}

\begin{figure}[h!]
  \centering
  \includegraphics[height=0.35\linewidth]{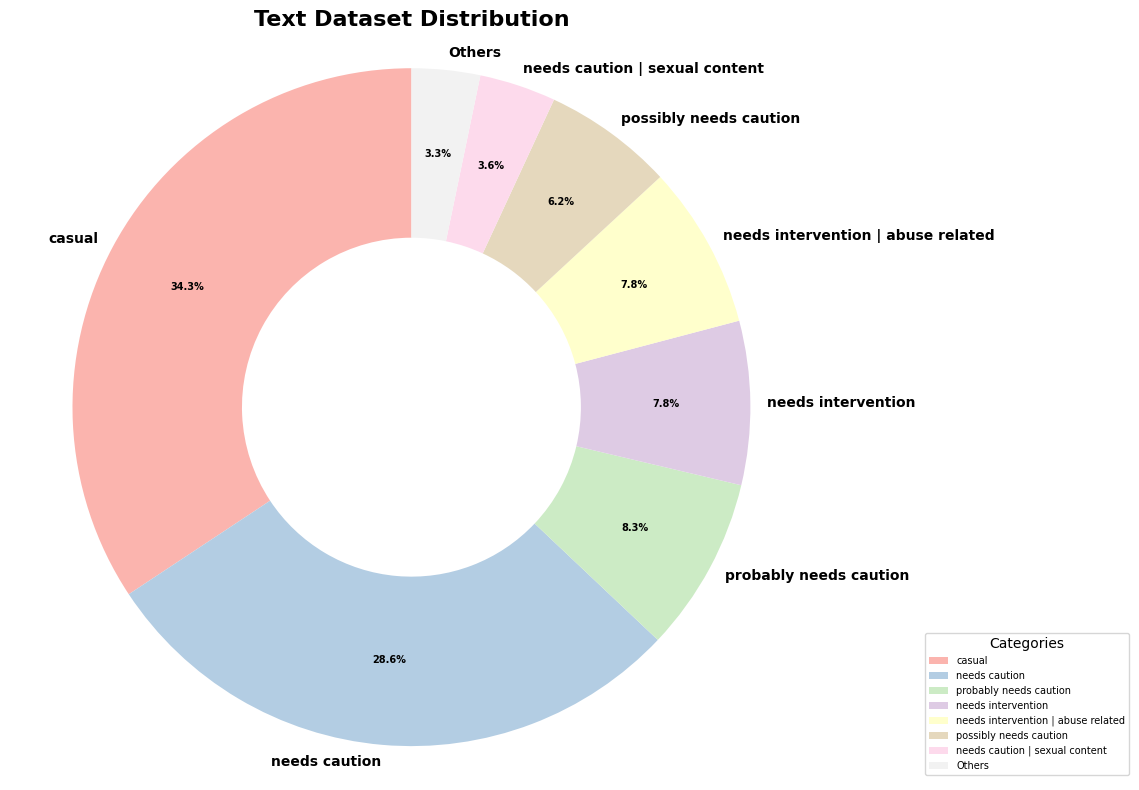}
  \caption{Text Dataset Distribution}
  \label{fig:textdist}
\end{figure}

\subsection{Dataset Statistics and Significance}

Our final dataset comprises:
- 558,958 text instruction-response pairs
- 83,625 image instruction-response pairs, a subset of the total combined image and text datasets described in \ref{tab:data_sources} selected at random.

In total, we processed 83,625 images using the GPT4V model and 558,958 text inputs using GPT4, to generate instruction-response pairs. The detailed distribution of our image-based instruction data is presented in Figure \ref{fig:imagedist} and our text-based instruction data is presented in Figure \ref{fig:textdist}.



\section{Model Fine-Tuning}

\subsection{Mod-RWKV}

We introduce a novel model called \textbf{Moderation-RWKV} (Mod-RWKV), a variant of the version 5 RWKV[\citealp[]{peng2023rwkv}] designed specifically for content moderation tasks. This model containing 3 billion parameters is optimized using our tailored text instruction dataset (see Table \ref{tab:instruction_examples}). The Mod-RWKV model, rooted in RNN architecture [\citealp[]{cho2014learning}], achieves GPT-level LLM performance and supports direct training comparable to a GPT transformer, allowing for parallelization. This attribute significantly enhances its inference speed and training efficiency, making Mod-RWKV an effective tool for large-scale moderation challenges.

\begin{itemize}    
    \item \textbf{Training Configuration:}
    \begin{itemize}
        \item \textbf{Learning Rate:} Adaptive, starting at 0.0006, reducing to 0.0004, with Adam optimization (beta1 = 0.9, beta2 = 0.99).
        \item \textbf{Gradient Checkpointing:} Enabled for improved memory efficiency.
        \item \textbf{Low-Rank Adaptation (LoRA) [\citealp[]{hu2021lora}] Parameters:} $r = 8$ and $\alpha = 16$, with a LoRA dropout rate of 0.01.
        \item \textbf{Weight Decay:} Set at 0.01 to help prevent overfitting.
    \end{itemize}

    \item \textbf{Training Environment and Hardware:}
    \begin{itemize}
        \item \textbf{Duration:} The training duration was around 1 day, 16 hours.
        \item \textbf{Hardware:} Utilized 24 CPU cores and 4 NVIDIA L4 GPUs.
        \item \textbf{Datasets:} Included Civil Comments[\citealp[]{DBLP:journals/corr/abs-1903-04561}] and OIG Moderation[\citealp[]{ontocord}] merged into single dataset format (see Table \ref{tab:instruction_examples}).
    \end{itemize}
\end{itemize}

\subsection{Mod-LLaVA}

\textbf{Moderation-LLaVA} (Mod-LLaVA) an adaptation of the LLaVA[\citealp[]{liu2023improved}] version 1.5 framework tailored for content moderation. This model is available in two configurations with 7 billion and 13 billion parameters and is fine-tuned with our specialized image instruction dataset (see Figure \ref{fig:exampleconversationwithrwkvandvisualrwkv}). The Mod-LLaVA, leveraging the LLaVA architecture, is designed to address the unique challenges of image-based content moderation.

\begin{itemize}
    \item \textbf{Dataset Balancing and Preprocessing:} Our initial dataset predominantly consisted of NSFW images, which could lead to biased model performance. To address this imbalance, we strategically removed a portion of the NSFW content, ensuring a more equitable distribution across categories. This balancing act was crucial in creating a dataset that better represents the diverse content encountered in real-world content moderation scenarios.
    
    \item \textbf{Vision Processing:} LLaVa integrates a pre-trained CLIP [\citealp[]{radford2021learning}] ViT-L/14 visual encoder [\citealp[]{dosovitskiy2020image}] with the Vicuna language model [\citealp[]{vicuna2023}] through a straightforward projection matrix. We initiated with a pre-trained model and subsequently tailored it to our specific dataset through the following stages:
        \begin{itemize}
            \item Stage 1: Feature Alignment Pre-training. At this stage, adjustments are made solely to the projection matrix, leveraging a segment of the CC3M[\citealp[]{changpinyo2021conceptual}] dataset to ensure alignment of features.
            \item Stage 2: Comprehensive End-to-End Fine-tuning:
            \begin{itemize}
                \item Visual Chat: In this phase, LLaVA undergoes fine-tuning with our custom-created multimodal dataset tailored for everyday conversational applications, enhancing its responsiveness and interaction in visual chat scenarios.
                \item Science QA [\citealp[]{lu2022learn}]: Here, LLaVA is fine-tuned using a specialized multimodal dataset focused on scientific queries and answers, aiming to bolster its performance in the science domain.
                \item Image Instruction: Fine-tuning LLaVA further, we use a bespoke dataset comprising images relevant to content moderation, refining the model's capability to process and understand image-based instructions (see Figure \ref{fig:exampleconversationwithrwkvandvisualrwkv}).
            \end{itemize}
        \end{itemize}
    \item \textbf{Training Configuration:} The training was executed using mixed precision (bf16) for a single epoch. Specific batch sizes were set for training and evaluation, with a gradient accumulation setting of 1. We implemented a step-based saving strategy and did not perform intermediate evaluations.
    
    \item \textbf{Learning Rate:} An initial learning rate of $2 \times 10^{-5}$ was used, with a cosine learning rate scheduler and a warmup ratio of 0.03. 
    
    \item \textbf{Logging and Memory Optimization:} Step-by-step logging was conducted, and gradient checkpointing was activated to optimize memory usage. Lazy preprocessing was also utilized.
    
    \item \textbf{Computational Resources:}The training was conducted on 8 A100 SXM 80GB GPUs, supported by 256 vCPUs and 2002 GB of RAM, incurring a cost of \$18 per hour. Training the model with 7 billion parameters required 4 hours, amounting to a total cost of \$72. The 13 billion parameter model took 8 hours to train, resulting in a total cost of \$144.
\end{itemize}

\subsection{Mod-VisualRWKV}
\textbf{Moderation-VisualRWKV} (Mod-VisualRWKV) is a specialized version of the VisualRWKV[\citealp[]{duan2024visionrwkv}] model version 5, crafted with a focus on image-based content moderation. This adaptation, equipped with 3 billion parameters, undergoes fine-tuning with the image instruction dataset (see Figure \ref{fig:exampleconversationwithrwkvandvisualrwkv}) specifically developed for this purpose. The Mod-VisualRWKV, grounded in the same foundational architecture as VisualRWKV, is tailored to meet the unique demands of visual content moderation, offering robust performance and efficient parallelizable training capabilities and as a powerful solution for moderating visual content at scale.

\begin{itemize}
    \item \textbf{Dataset Balancing and Preprocessing:} Our initial dataset predominantly consisted of NSFW images, which could lead to biased model performance. To address this imbalance, we strategically removed a portion of the NSFW content, ensuring a more equitable distribution across categories. This balancing act was crucial in creating a dataset that better represents the diverse content encountered in real-world content moderation scenarios.
    
    \item \textbf{Vision Processing:} For vision processing, we employed the CLIP ViT-L/14 vision tower [\citealp[]{dosovitskiy2020image},\citealp[]{radford2021learning}]. This component processes images and converts them into embeddings that are then projected to match the RWKV's embedding size, allowing for seamless integration with textual data.
    
    \item \textbf{Model Initialization:} The VisualRWKV model initializes by creating an RWKV instance and loading pretrained weights if available. It then establishes the vision processing component and a projection layer to align the dimensions of visual embeddings with the RWKV model's embeddings.
    
    \item \textbf{Freezing Layers:} The model provides methods to freeze either the RWKV layers or the projection layer, which is beneficial during fine-tuning to maintain certain learned representations.
    
    \item \textbf{Forward Pass:} In the forward pass, image and text data are processed to generate embeddings, which are then passed through the RWKV model to produce outputs, such as logits for classification tasks.
    
    \item \textbf{Training Configuration:} The training was executed using mixed precision (bf16) on NVIDIA A100 GPUs for a single epoch. A step-based saving strategy was implemented without performing intermediate evaluations.
    
    \item \textbf{Logging and Memory Optimization:} Step-by-step logging was conducted, and gradient checkpointing was activated to optimize memory usage. Lazy preprocessing was also utilized.
    
    \item \textbf{Computational Resources:} The training utilized 8 A100 SXM 80GB GPUs, with 256 vCPUs and 2002 GB RAM, at a cost of \$18 per hour. The training duration was 3 hours.
\end{itemize}

\begin{figure}[h!]
\def\arraystretch{1.2}
\centering
\begin{tabular}{|c|c|}
\hline
\text{Model} & \text{Accuracy (\%)} \\
\hline
\text{RWKV 3B} & 58.7\% \\
\text{Mod-RWKV 3B} & \textbf{66.9\%} \\
\hline
\text{VisualRWKV 3B} & 59.1\% \\
\text{Mod-VisualRWKV 3B} & \textbf{84.8\%} \\
\hline
\text{LLaVA 7B} & 83.2\% \\
\text{Mod-LLaVA 7B} & \textbf{86.8\%} \\
\hline
\text{LLaVA 13B} & 82.4\% \\
\text{Mod-LLaVA 13B} & \textbf{87.3\%} \\
\hline
\end{tabular}
    \captionof{table}{Accuracies for various models based on 1000 sample evaluations}
    \label{tab:accuracy_total}
\end{figure}
\section{Experiments and Evaluations}

In this comprehensive evaluation, we assess the performance of our proposed models against the benchmarks established by the ToxiGen[\citealp[]{hartvigsen2022toxigen}] and NSFW images from Porn Lab Dataset \cite{porn_lab_dataset}, which are pivotal in determining the efficacy of content moderation models. Our evaluation primarily hinges on accuracy, gauging the proportion of instances where the model's classifications align with the expected outcomes.
For textual content (text evaluations), we employ an LLM-as-judge framework. Here, GPT-4 [\citealp[]{openai2024gpt4}] serves as an adjudicator, assessing the correctness of model outputs against textual inputs drawn randomly from the ToxiGen dataset [\citealp[]{hartvigsen2022toxigen}]. This process involves presenting GPT-4 with a series of instructions and input messages, whereupon it determines the appropriateness of the model-generated responses. This method not only tests the model's understanding of context and nuance in textual data but also its alignment with human-like judgment standards. Parallel to our text evaluations, we implement an LLM-as-judge framework for images, utilizing GPT-4V's [\citealp[]{2023GPT4VisionSC}] capabilities to judge image-related outputs. In this setup, an image is selected from NSFW images found in Porn Lab Dataset \cite{porn_lab_dataset} and paired with random instructions and input messages. GPT-4V then assesses whether the model's response to the image input is accurate or not. This dual evaluation, encompassing both text and image assessments, enables us to gauge the model's multimodal understanding and its proficiency in interpreting and moderating visual content.

\section{Limitations and Future Work}
The dataset we utilized has notable shortcomings that necessitate further attention. A critical issue is the lack of image diversity, which is essential for a more comprehensive representation. Moreover, our dataset, like others, may have embedded biases, such as racial profiling. Limited access to GPUs and computational resources also posed challenges, restricting our capacity to develop an optimal model. Additionally, we encountered limitations with the BLIP[\citealp[]{li2022blip}] framework's performance. 

In our current work, we employed Supervised Fine-Tuning (SFT) \cite{gunel2020supervised} to adapt the RWKV model for content moderation tasks. While this approach yielded promising results, future iterations of our research should explore more advanced techniques such as Direct Preference Optimization (DPO) \cite{qi2024online} and Reinforcement Learning from Human Feedback (RLHF) \cite{huyen2023rlhf} . These methods could potentially lead to more robust and aligned models, better suited for the nuanced task of content moderation.

Improving the dataset's breadth and investing in more robust data preprocessing could significantly enhance our large language model's (LLM) effectiveness. The current instruction set's limitations, applicable to both text and images, could introduce biases, favoring certain instruction types. Expanding content moderation to encompass religious content poses another challenge, as our model lacks training in this area. 

Future initiatives will focus on diversifying the image dataset and considering the development of models like RWKV from the ground up instead of merely fine-tuning them. We also aim to adopt an output format akin to the OpenAI moderation API, providing a JSON file with categorized labels and scores. Ultimately, we plan to transcend the current instruction-based modeling approach, exploring alternative strategies for model development and enhancement, including the aforementioned DPO \cite{qi2024online} and RLHF \cite{huyen2023rlhf} techniques.
\section{Ethics}
Detecting hate speech is intricate, as simplifying it to basic logical rules might inadvertently embed deep biases, potentially leading to issues like erasure when specific in-group or identity terms trigger content flags. In the realm of handling a multimodal dataset with sensitive elements like violence and NSFW content, ethical diligence is crucial \cite{bengani2018controlling}. It's essential to implement stringent security protocols to prevent unauthorized data access and ensure confidentiality. Equally important is the need to tackle inherent biases in the dataset to foster fairness and inclusivity. Adopting ethical practices for data labeling and annotation \cite{gurav2019accessible}, along with clear explanations of algorithmic choices, is key to reducing biases and maintaining transparency. Establishing guidelines for the responsible application of content moderation models is vital to reduce potential harms and encourage ethical usage. Engaging with stakeholders and conducting regular evaluations of the dataset and model effectiveness are crucial steps to uphold ethical norms and promote continuous enhancement. By focusing on these ethical aspects, researchers can create content moderation models that not only effectively filter out damaging content but also adhere to high ethical standards, safeguarding individual rights and dignity.
\section{Conclusion}
This study focused on experimenting with content moderation models using a curated multimodal dataset. We conducted extensive tests on model architectures designed to detect and filter harmful content across text, images, audio, and video. Our experiments leveraged a diverse dataset covering various content types, including violence and NSFW material. These experiments not only advanced our understanding of effective content moderation techniques but also highlighted areas for future research. Continued experimentation on content moderation using LLMs could be beneficial. This ongoing innovation, driven by rigorous testing, better datasets, and context understanding models, will be key to developing more effective content moderation technologies and fostering safer online spaces.

\bibliographystyle{IEEEtranN}
\bibliography{main}  
\end{document}